\documentclass[runningheads]{llncs}
\usepackage[T1]{fontenc}
\usepackage{graphicx,verbatim}

\usepackage{multirow}
\usepackage{makecell}
\usepackage{booktabs}
\usepackage{colortbl}
\usepackage{xcolor}
\usepackage{mathabx}

\begin{document}
\title{Leveraging Spatial Context for Positive Pair Sampling in Histopathology Image Representation Learning}
\titlerunning{Spatial Context for Positive Pair Sampling in Histopathology}
% \begin{comment}  %% Removed for anonymized MICCAI submission
\author{
    Willmer Rafell Quinones Robles\inst{1} \and
    Sakonporn Noree\inst{1} \and
    Jongwoo Kim\inst{1} \and
    Young Sin Ko\inst{2} \and
    Bryan Wong\inst{1} \and
    Mun Yong Yi\inst{1}
}

\authorrunning{W. Quinones et al.}
\institute{Korea Advanced Institute of Science and Technology, Daejeon, South Korea  \\
\email{\{wrafell,sakonporn.n,gsds4885,bryan.wong,munyi\}@kaist.ac.kr}\and
Seegene Medical Foundation, Seoul, South Korea \\
\email{noteasy@mf.seegene.com}}

% \end{comment}

% \author{Anonymized Authors}  %% Added for anonymized MICCAI submission
% \authorrunning{Anonymized Author et al.}
% \institute{Anonymized Affiliations \\
%     \email{email@anonymized.com}}
  
\maketitle              % typeset the header of the contribution
\begin{abstract}
\label{sec:abstract}
Deep learning has achieved strong performance in cancer classification from whole-slide images (WSIs), but its reliance on large-scale expert annotations limits scalability. Self-supervised learning (SSL) alleviates this constraint by learning patch-level representations from unlabeled data; however, most methods rely on synthetic augmentations that ignore the spatial continuity of tissue. We propose a spatially coherent positive pair sampling strategy that improves patch-level representation learning by incorporating spatially adjacent patches as contextual positives. Our method is architecture-agnostic and integrates into standard joint-embedding SSL frameworks without architectural changes or triplet losses. Across multiple datasets and SSL frameworks, improvements in patch-level embeddings translate into consistent gains in slide-level classification with multiple instance learning and in patch-level linear probing, outperforming augmentation-only baselines. These results demonstrate that enforcing local spatial coherence yields more informative patch representations for computational pathology. Code is available at https://anonymous.4open.science/r/contextual-pairs-BCA4/

% Evaluated across multiple datasets and SSL frameworks using slide-level classification with multiple instance learning and patch-level linear probing, our approach consistently outperforms augmentation-only baselines. The results demonstrate that leveraging local spatial coherence improves representation learning in computational pathology. 


% Authors must provide keywords and are not allowed to remove this Keyword section.

\end{abstract}
\section{Introduction}
\label{sec:intro}

Deep learning has substantially improved cancer diagnosis from whole-slide images (WSIs) across classification and segmentation tasks. However, these advances depend on large-scale expert annotations, which are costly and difficult to scale. To reduce the annotation burden, multiple instance learning (MIL) \cite{Courtiol2018c} has emerged as the de facto method \cite{abmil,dsmil,dtfdmil,transmil}. Particularly, MIL methods often use patch-level encoders trained in a self-supervised learning (SSL) manner. Recent work on pathology foundation models highlights the effectiveness of SSL objectives as patch encoders \cite{Cubuk2020,hibou,conch,kang2023}.

    \begin{figure*}[ht]
    \centering
    \includegraphics[width=\linewidth]{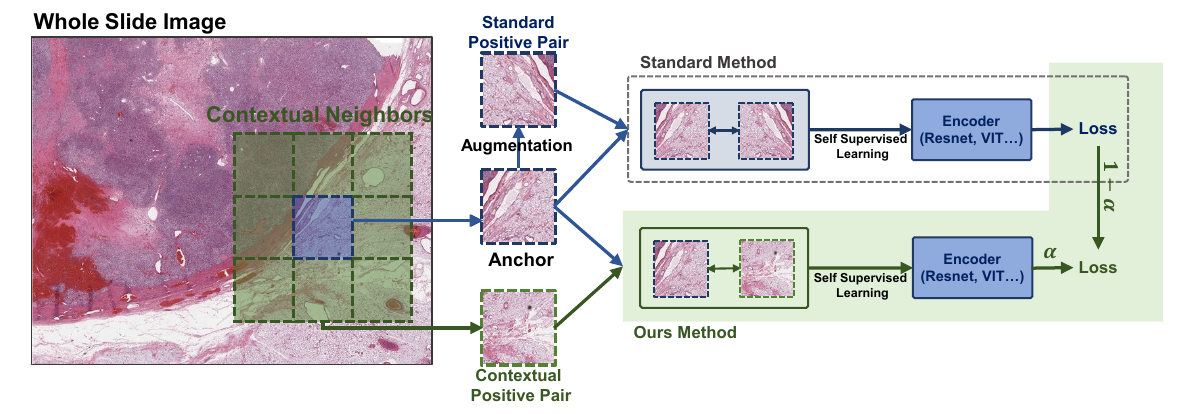}
    \caption{Standard SSL forms positives via augmentations of a single anchor patch. Our method additionally samples spatially adjacent patches within a predefined neighborhood to form contextual positive pairs (green).}
    \label{fig:overview}
    \end{figure*}

However, current SSL frameworks \cite{simclr,ssl_survey,byol,vicreg,dinov2} rely primarily on \textit{augmentation-based invariance}, where positive pairs are generated by applying synthetic transformations (e.g., cropping, color jittering) to a single isolated patch. While this encourages invariance to appearance changes, it neglects a defining property of WSIs: strong spatial continuity of tissue morphology. A patch’s biological identity is tightly coupled with its local neighborhood, yet standard SSL either ignores adjacent patches or treats them as unrelated samples. Consequently, patch-level representations may lack sensitivity to local structural coherence.

To bridge this gap, we propose a \textbf{spatially coherent positive pair sampling strategy} tailored to histopathology. In addition to standard augmentations, we incorporate spatially adjacent patches as contextual positives, leveraging the inherent supervision embedded in tissue architecture (Figure~\ref{fig:overview}). Our method is modular and architecture-agnostic, integrating directly into joint-embedding SSL frameworks without triplet networks or structural changes.

Prior work has explored spatial proximity as supervision \cite{tile2vec,Pantazis2021,simtriplet}. In particular, SimTriplet \cite{simtriplet} exploits adjacent patches in WSIs but relies on a triplet architecture and custom loss to model intra- and inter-sample relations. In contrast, we introduce a flexible sampling strategy that blends augmentation-based and spatial positives within standard SSL objectives, providing a lightweight improvement to patch-level representation learning.

We evaluate our method across multiple datasets and backbones. Rather than competing with large-scale foundation models, we present a biologically grounded, lightweight enhancement to SSL that improves representation quality for digital pathology with minimal overhead.

\section{Methodology}
\label{sec:methods}
\subsection{Contextual Positive Pair Sampling}
We introduce a spatially-aware positive pair sampling strategy that leverages the local structural coherence of tissue morphology. The key idea is to incorporate spatially adjacent patches as contextual signals, augmenting standard augmentation-based sampling. An overview of this strategy is shown in Figure~\ref{fig:overview}. Note that our method is modular and compatible with a variety of SSL approaches.

Let $U^P$ denote a set of $M$ \textit{pivot patches} sampled from the WSIs in the training set. For each pivot patch $p^i_j \in U^P$, we identify a set of $K$ \textit{neighbor patches} $U^N = \{p^i_k\}$ within the same WSI, constrained by a maximum Chebyshev distance $d$. The hyperparameter $d$ defines the spatial neighborhood radius and directly controls the degree of contextual granularity: small $d$ enforces highly local morphological consistency, while larger $d$ captures broader tissue context. We adopt the Chebyshev metric due to its natural alignment with the grid structure of WSIs, where patches are spatially indexed.

During training, we construct two types of positive pairs:
\begin{enumerate}
    \item \textbf{Standard augmentation-based pair}: Two augmented views of the same patch, $v_{1,j} = t_1(p^i_j)$ and $v_{2,j} = t_2(p^i_j)$, where $t_1, t_2 \sim T$ are sampled transformations.
    \item \textbf{Contextual pair}: A neighbor patch $p^i_k \in U^N$ is randomly selected within distance $d$, and both $p^i_j$ and $p^i_k$ are augmented using the same transformation $t_1$, producing $v_{1,j} = t_1(p^i_j)$ and $v_{1,k} = t_1(p^i_k)$.
\end{enumerate}

The combined loss function is defined as:
\begin{equation}
    L' = \alpha \cdot L(v_{1,j}, v_{1,k}) + (1 - \alpha) \cdot L(v_{1,j}, v_{2,j})
    \label{eq:main_loss}
\end{equation}
where $L(\cdot, \cdot)$ denotes the loss function from the SSL framework in use (e.g., redundancy reduction for Barlow Twins or BYOL). The weight $\alpha \in [0,1]$ balances contextual similarity and transformation invariance. This formulation avoids the need for negative pairs and is particularly well-suited for WSI data, where distant patches may still belong to the same class.

\subsection{False Positive Pairs}  
\begin{figure}[ht]
\centering
\includegraphics[width=0.4\linewidth]{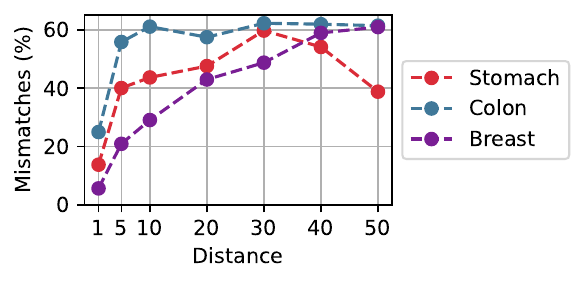}
\caption{Percentage of mismatches as a function of Chebyshev distance.}
\label{fig:mismatch_stats}
\end{figure}

Sampling neighboring patches by spatial distance introduces the risk of false positive pairs (mismatches), particularly when adjacent regions belong to different tissue types or classes. To quantify this effect, we measured the mismatch rate as a function of sampling distance (Figure~\ref{fig:mismatch_stats}), restricting the analysis to non-benign WSIs since benign slides contain only benign patches.

Mismatch rates increase with distance, indicating that patches from farther regions are more likely to differ semantically. Based on this observation, we restrict contextual positives to immediate neighbors (maximum Chebyshev distance of 1), randomly selecting one neighbor per pivot during training. We set $\alpha = 0.5$ to balance contextual similarity and augmentation-based invariance. The influence of distance thresholds and $\alpha$ is further analyzed in ablation studies, along with a variant that samples neighbors without distance constraints.

\section{Experiments and Results}
\label{sec:experiments}
\subsection{Datasets}
% \paragraph{Public Datasets}  
We evaluated our method on four datasets: two publicly available and two private ones. We used Camelyon16 \cite{Bejnordi2017} and TCGA-NSCLC as the publicly available datasets. The \textbf{Camelyon16} (Breast) dataset consists of 399 whole slide images (WSIs) of lymph node sections, including 257 normal slides and 142 slides with metastatic regions. It is divided into a training set of 270 slides and a test set of 129 slides. We further partitioned the training set into 80 normal and 80 malignant slides for training, with an additional 80 normal and 30 malignant slides reserved for validation. \textbf{TCGA-NSCLC} (Lung) dataset comprises 1,046 WSIs, with 836 slides used for training and 210 for testing, spanning two lung cancer subtypes: 534 Lung Adenocarcinoma (LUAD) and 512 Lung Squamous Cell Carcinoma (LUSC). On both datasets, we extracted non-overlapping 256x256 pixel tiles from the tissue regions at 20x magnification.

% \input{tables/data}

% \paragraph{Private Datasets}
% We evaluated our method in two private datasets acquired from a medical foundation: (1) 600 stomach WSIs and (2) 600 colorectal WSIs. Each WSI contains 2 to 4 slices from the same tissue sample. Among the slides, and for each dataset, 200 were classified as normal/benign slides (N), 200 were classified as tissues in dysplasia (D), and 200 were classified as cancerous/malignant (M). The Institutional Review Board (Approval \#SMF-IRB-2020–007) of the organization and the Institutional Review Board (Approval \#KAIST-IRB-22-334, \#KH2020-116) of Korea Advanced Institute of Science and Technology (KAIST) have both granted approval for research use. The WSIs were stained with hematoxylin and eosin (H\&E), digitized as MIRAX files with a sensor resolution of 200x, using a Panoramic Flash 250 III scanner, and were examined and graded by two pathologists from the same organization. All patient records were anonymized. Moreover, for each class, we randomly split the WSIs into 140 slides for training, 30 slides for validation, and 30 slides for testing. Using a sliding window approach, we extracted 256x256 pixel patches at 20x magnification with the OpenSlide Python library \cite{goode2013}.
We further evaluated our method on two private datasets obtained from a collaborating medical institution: (1) 600 WSIs of stomach tissue and (2) 600 WSIs of colorectal tissue. Each WSI comprises 2 to 4 sections from the same specimen. For each dataset, the slides were evenly categorized into normal/benign (N), dysplastic (D), and malignant (M) groups, with 200 slides per class. For each class, we randomly split the slides into 140 for training, 30 for validation, and 30 for testing. Patches of size 256$\times$256 pixels were extracted at 20$\times$ magnification using a sliding window approach implemented with the OpenSlide Python library \cite{goode2013}.

Research use of the private datasets was approved by the Institutional Review Boards of Institution A (Approval \#InstA-IRB-2020–007) and Institution B (Approval \#InstB-IRB-22-334, \#InstB-2020-116). All WSIs were stained with hematoxylin and eosin (H\&E), digitized in MIRAX format at a resolution of 200x using a Panoramic Flash 250 III scanner, and independently reviewed by two board-certified pathologists affiliated with Institution A. Patient data were fully anonymized prior to analysis.

\subsection{Implementation Details}

% All experiments were implemented in PyTorch~\cite{paszke2019} on an RTX 3080 Ti (32 GB). For patch-level representation learning, we evaluated \textbf{Barlow Twins}~\cite{barlow_twins}, \textbf{BYOL}~\cite{byol}, \textbf{VICReg}~\cite{vicreg}, and \textbf{DINOv2}~\cite{dinov2} using ResNet-18~\cite{He2016} (ViT-Tiny for DINOv2). Barlow Twins, BYOL, and VICReg were trained with LARS~\cite{lars_optimization} (lr 0.2/0.0048, weight decay $1\times10^{-6}$) for 100 epochs (batch size 512); DINOv2 used AdamW (lr $5\times10^{-4}$, weight decay 0.04) with cosine scheduling and teacher temperature warm-up ($0.02{\to}0.04$, 15 epochs) for 50 epochs (batch size 128). Augmentations followed~\cite{simclr,kang2023,dinov2}; local crops omitted; inputs were ImageNet-normalized.

% For slide-level classification, frozen patch embeddings were aggregated via ABMIL~\cite{abmil} trained with Adam (lr $1\times10^{-3}$, weight decay $1\times10^{-2}$, 100 epochs). Linear probing on frozen encoders (5k/2.5k/2.5k train/val/test patches per class) used SGD (ResNet-18, lr 0.1, cosine) or AdamW (ViT-Tiny, lr $3\times10^{-4}$, weight decay 0.01) for 50 epochs. A supervised baseline fine-tuned the same backbones end-to-end. TCGA was excluded from patch-level experiments due to missing annotations. Performance was assessed via accuracy and AUC-ROC, averaged over five runs (seeds 0--4).

All experiments were implemented in PyTorch~\cite{paszke2019} on an RTX 3080 Ti (32\,GB). For patch-level representation, we evaluated \textbf{Barlow Twins}~\cite{barlow_twins}, \textbf{BYOL}~\cite{byol}, \textbf{VICReg}~\cite{vicreg}, and \textbf{DINOv2}~\cite{dinov2} using ResNet-18~\cite{He2016} (ViT-Tiny for DINOv2). Barlow Twins, BYOL, and VICReg were trained with LARS~\cite{lars_optimization} for 100 epochs (batch size 512), while DINOv2 used AdamW with cosine scheduling and teacher temperature warm-up for 50 epochs (batch size 128). Augmentations followed prior work~\cite{simclr,kang2023,dinov2}, and inputs were ImageNet-normalized.

For slide-level classification, frozen patch embeddings were aggregated using ABMIL~\cite{abmil} trained with Adam for 100 epochs. Linear probing on frozen encoders (5k/2.5k/2.5k train/val/test patches per class) was conducted for 50 epochs using SGD (ResNet-18) or AdamW (ViT-Tiny). A supervised baseline fine-tuned the same backbones end-to-end. TCGA was excluded from patch-level experiments due to missing annotations. Performance was evaluated using accuracy and AUC-ROC, averaged over five runs (seeds 0--4).

\subsection{Slide-level Classification}

    \begin{table*}
    \fontsize{8}{12}\selectfont
    \caption{Slide-level performance across datasets (averaged over 5 runs). Gray cells indicate contextual approaches: ($\infty$) for unrestricted distance sampling and ($1$) for Chebyshev distance of $1$. \textit{BT} denotes Barlow Twins. \textbf{Bold} indicates highest value.}
    \label{tab:ssl_comparison}
    \centering
    \begin{tabular}{ccc>{\columncolor{gray!20}}c>{\columncolor{gray!40}}c|c>{\columncolor{gray!20}}c>{\columncolor{gray!40}}c}
    \hline
    \multirow{2}{*}{Dataset} & \multirow{2}{*}{SSL} & \multicolumn{3}{c}{Accuracy} & \multicolumn{3}{c}{AUROC} \\
    % \cline{3-8}
     &  & Standard & \cellcolor{gray!20}Context($\infty$) & \cellcolor{gray!40}Context($1$) & Standard & \cellcolor{gray!20}Context($\infty$) & \cellcolor{gray!40}Context($1$) \\
    \hline
    \multirow{4}{*}{\thead{Private\\ (Stomach)}} & BT & 0.827 & 0.836 & \textbf{0.893} & 0.941 & 0.947 & \textbf{0.966} \\
     & BYOL & 0.813 & 0.833 & \textbf{0.902} & 0.954 & 0.964 & \textbf{0.972} \\
     & VICReg & 0.831 & 0.816 & \textbf{0.893} & 0.947 & 0.945 & \textbf{0.960} \\
     & DINOv2 & 0.867 & 0.893 & \textbf{0.896} & 0.962 & 0.967 & \textbf{0.972} \\
    \hline
    \multirow{4}{*}{\thead{Private\\ (Colon)}} & BT & 0.820 & 0.844 & \textbf{0.891} & 0.926 & 0.945 & \textbf{0.971} \\
     & BYOL & 0.802 & 0.756 & \textbf{0.916} & 0.922 & 0.930 & \textbf{0.987} \\
     & VICReg & 0.807 & 0.809 & \textbf{0.916} & 0.926 & 0.931 & \textbf{0.981} \\
     & DINOv2 & 0.789 & 0.841 & \textbf{0.867} & 0.919 & \textbf{0.965} & 0.954 \\
    \hline
    \multirow{4}{*}{\thead{Camelyon16\\ (Breast)}} & BT & 0.710 & 0.710 & \textbf{0.763} & 0.656 & 0.658 & \textbf{0.722} \\
     & BYOL & 0.757 & 0.767 & \textbf{0.798} & 0.751 & 0.738 & \textbf{0.753} \\
     & VICReg & 0.757 & 0.716 & \textbf{0.798} & 0.726 & 0.705 & \textbf{0.777} \\
     & DINOv2 & 0.788 & 0.817 & \textbf{0.835} & 0.769 & 0.799 & \textbf{0.863} \\
    \hline
    \multirow{4}{*}{\thead{TCGA\\ (Lung)}} & BT & 0.773 & 0.771 & \textbf{0.825} & 0.839 & 0.835 & \textbf{0.909} \\
     & BYOL & 0.763 & 0.760 & \textbf{0.849} & 0.847 & 0.844 & \textbf{0.920} \\
     & VICReg & 0.770 & 0.769 & \textbf{0.857} & 0.846 & 0.846 & \textbf{0.914} \\
     & DINOv2 & 0.701 & 0.737 & \textbf{0.833} & 0.785 & 0.75 & \textbf{0.893} \\
    \hline
    \end{tabular}
    \end{table*}

Table~\ref{tab:ssl_comparison} compares standard positive pair sampling with two contextual variants. Across all datasets and SSL methods, Context($1$) consistently outperforms the standard approach in both accuracy and AUROC, demonstrating the benefit of leveraging immediate spatial context. In contrast, Context($\infty$) yields mixed results, suggesting that unconstrained sampling may introduce less informative or noisy pairs. The magnitude of improvement with Context($1$) varies by dataset: for example, the Colon dataset shows gains exceeding 10\% in accuracy using \textit{VICReg}, while the Breast dataset exhibits more moderate improvements (around 4\%). Despite this variability, Context($1$) consistently provides the best overall performance, indicating that enforcing local spatial coherence during patch-level SSL leads to more robust representations for slide-level classification.

\subsection{Patch-level Embedding Visualization}
We visualize patch-level embeddings using t-SNE \cite{tsne} and report Normalized Mutual Information (NMI) \cite{strehl2002cluster}, showing results for VICReg and DINOv2 on the Private (stomach) dataset due to space constraints. Although patch-level evaluation is inherently noisy because annotations are provided at coarse granularity \cite{Ashraf2022,Park2022,Brunye2017}, contextual positive sampling yields clearer class separation and more compact clusters than augmentation-only training, reflected in higher NMI scores (VICReg: 0.018 $\rightarrow$ 0.231; DINOv2: 0.011 $\rightarrow$ 0.156).

    \begin{figure}[ht]
    \centering
    \includegraphics[width=\linewidth]{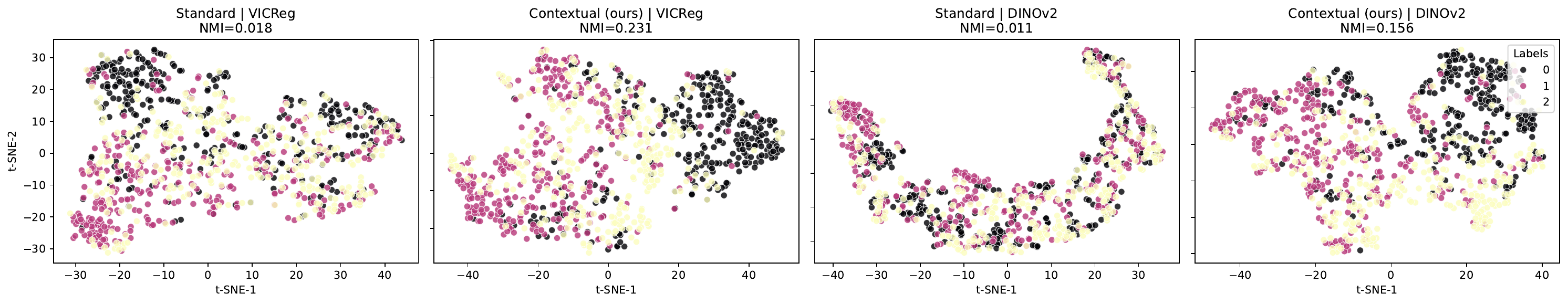}
    \caption{t-SNE of patch embeddings on the Private (stomach) dataset using VICReg (first 2 subplots) and DINOv2 (last 2 subplots). Contextual sampling (second and fourth subplots) yields higher NMI than standard training.}
    \label{fig:tsne-stomach}
    \end{figure}

\subsection{Linear Probing Evaluation}

    \begin{table*}
    \fontsize{8}{12}\selectfont
    \caption{
    Patch-level performance (via linear probing) across datasets (averaged over 5 runs). TCGA excluded (no patch labels). \textit{Standard} refers to SSL pretraining without contextual pairing, while \textit{Context(1)} refers to our method. \textit{Supervised} indicates fully supervised training; the first value corresponds to a ResNet-18, and the second (in the DINOv2 row) to a ViT-Tiny.}
    \label{tab:linear_probing_results}
    \centering
    \begin{tabular}{ccc>{\columncolor{gray!20}}c>{\columncolor{gray!40}}c|c>{\columncolor{gray!20}}c>{\columncolor{gray!40}}c}
    \hline
    \multirow{2}{*}{Dataset} & \multirow{2}{*}{SSL} & \multicolumn{3}{c}{Accuracy} & \multicolumn{3}{c}{AUROC} \\
     &  & Supervised & Standard & Context($1$) & Supervised & Standard & Context($1$) \\
    \hline
    \multirow{4}{*}{\thead{Private\\ (Stomach)}} 
     & BT      & \multirow{3}{*}{0.828} & 0.739 & 0.783 & \multirow{3}{*}{0.942} & 0.912 & 0.928 \\
     & BYOL    &                        & 0.774 & 0.809 &                        & 0.913 & 0.938 \\
     & VICReg  &                        & 0.803 & 0.821 &                        & 0.933 & 0.944 \\
     & DINOv2  & 0.830                 & 0.750 & 0.762 & 0.944                 & 0.896 & 0.904 \\
    \hline
    \multirow{4}{*}{\thead{Private\\ (Colon)}} 
     & BT      & \multirow{3}{*}{0.818} & 0.840 & 0.827 & \multirow{3}{*}{0.943} & 0.939 & 0.935 \\
     & BYOL    &                        & 0.861 & 0.885 &                        & 0.950 & 0.963 \\
     & VICReg  &                        & 0.859 & 0.862 &                        & 0.950 & 0.952 \\
     & DINOv2  & 0.878                 & 0.797 & 0.858 & 0.962                 & 0.921 & 0.945 \\
    \hline
    \multirow{4}{*}{\thead{Camelyon16\\ (Breast)}} 
     & BT      & \multirow{3}{*}{0.885} & 0.764 & 0.756 & \multirow{3}{*}{0.947} & 0.831 & 0.817 \\
     & BYOL    &                        & 0.777 & 0.782 &                        & 0.834 & 0.821 \\
     & VICReg  &                        & 0.792 & 0.798 &                        & 0.860 & 0.877 \\
     & DINOv2  & 0.910                 & 0.797 & 0.858 & 0.964                 & 0.921 & 0.945 \\
    \hline
    \end{tabular}
    \end{table*}

To further evaluate patch-level representation quality, we conducted linear probing using both ResNet-18 and ViT-Tiny backbones (Table~\ref{tab:linear_probing_results}). The results show that incorporating spatially contextual positives (\textit{Context(1)}) consistently improves performance over standard SSL across most datasets and methods. Notably, our approach nearly closes the gap with supervised training and, in some cases (e.g., BYOL and VICReg on Colon), surpasses it.

\subsection{Ablation Studies}    

    \begin{figure}[ht]
    \centering
    \includegraphics[width=\linewidth]{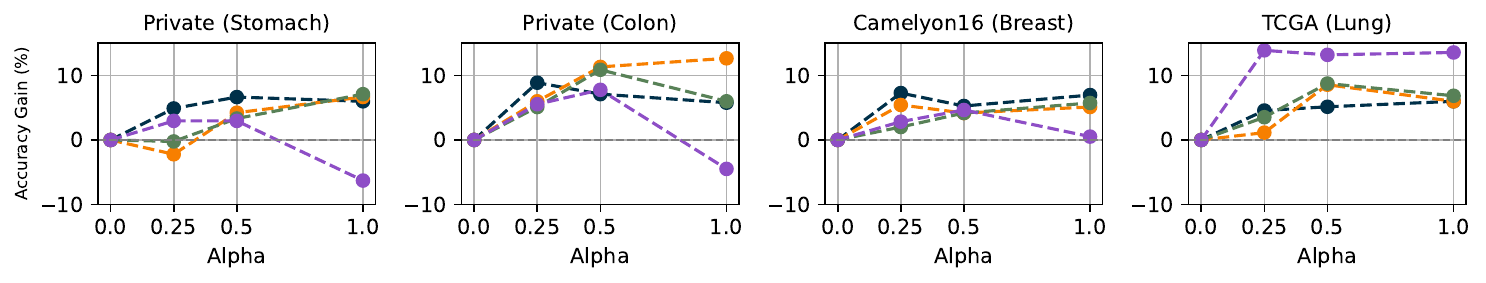}
    \caption{Accuracy gain (\%) relative to the baseline (i.e., without contextual information). Results are averaged over five runs. Dark blue, orange, green, and purple lines correspond to Barlow Twins (BT), BYOL, VICReg, and DINOv2, respectively.}
    \label{fig:ablation_alpha}
    \end{figure}

    \begin{figure}[ht]
    \centering
    \includegraphics[width=\linewidth]{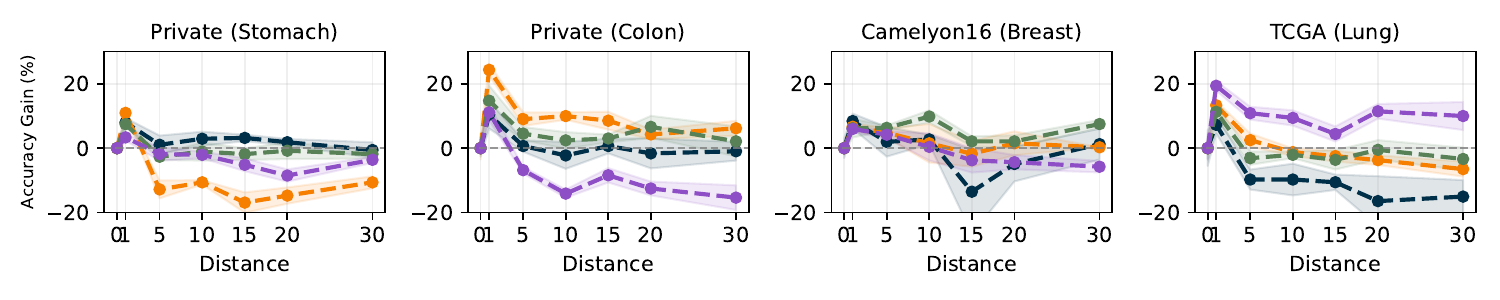}
    \caption{Effect of neighboring distance on slide-level classification accuracy measured via accuracy gain (\%). Error regions represent the standard deviation. Dark blue, orange, green, and purple lines correspond to Barlow Twins (BT), BYOL, VICReg, and DINOv2, respectively.}
    \label{fig:ablation_proximity}
    \end{figure}
    
\subsubsection{Proportion of Contextual Similarity ($\alpha$)}  

Our loss balances contextual similarity and transformation invariance through the mixing parameter $\alpha$. Figure~\ref{fig:ablation_alpha} shows that incorporating contextual signals ($\alpha > 0$) generally improves accuracy across datasets and methods. The best performance typically occurs at intermediate values ($\alpha = 0.25$ or $0.5$), while relying solely on contextual pairs ($\alpha = 1.0$) often degrades accuracy. This suggests that contextual cues alone may lack sufficient diversity, especially for transformer-based models. \textit{VICReg} (shown in green) shows stable improvements, while \textit{BYOL} (orange) is more sensitive to $\alpha$. \textit{BT} (dark blue) performs well at moderate $\alpha$ values, and \textit{DINOv2} (purple) benefits from some contextual information but drops sharply when overused.

\subsubsection{Effect of Neighboring Distance ($d$)}
    
We investigated how the spatial distance between pivot and neighbor patches affects the performance of self-supervised learning (SSL) in slide-level classification (Figure~\ref{fig:ablation_proximity}). From a pool of 100,000 pivot patches, we sampled neighbors within defined distance thresholds to capture both local and broader tissue context. Accuracy gains are highest when neighbor patches are sampled from short distances, with performance peaking consistently at distance 1 across all datasets and SSL methods. Beyond this point, accuracy generally declines, particularly at distances greater than 5, likely due to reduced contextual coherence and increased chances of mismatched or noisy pairs. The effect is consistent across SSL methods, although the extent of gain varies by dataset and method. Nonetheless, our results suggest that immediate spatial proximity (i.e., $distance = 1$) offers the most reliable and biologically meaningful context for representation learning.

\section{Discussion}
\label{sec:discussion}
We introduced a spatial context-driven positive pair sampling strategy for self-supervised learning (SSL) in whole-slide images (WSIs) that leverages tissue morphological continuity to improve patch-level representations. By incorporating spatially adjacent patches as positives, our method extends beyond augmentation invariance to encode local structural coherence. Enforcing immediate proximity (Chebyshev distance = 1) consistently improved slide-level classification and patch-level linear probing across datasets and SSL frameworks, with gains up to 4--8\%. Unlike previous work, which requires a dedicated triplet architecture and custom loss, our approach is architecture-agnostic and integrates seamlessly into joint-embedding frameworks via a unified weighted objective, enabling broad applicability without structural modification. While experiments were limited to a single magnification, fixed patch size, one backbone per framework, and a fixed distance threshold, the results demonstrate that embedding spatial structure into positive pair design enhances representation quality. Future work includes adaptive sampling strategies and extending spatially informed representations beyond classification to tasks such as retrieval and clustering.

\bibliographystyle{splncs04}
\bibliography{references}

\end{document}